\documentclass{article}

\newcommand{\IsPreprint}{1}
\usepackage{amsmath}
\usepackage{booktabs}
\usepackage[inline]{enumitem}
\usepackage{graphics}
\usepackage{graphicx}
\usepackage{hyperref}
\usepackage[utf8]{inputenc}
\usepackage{multirow}
\usepackage{pgfplots}
\usepackage[all=normal, paragraphs=tight, floats=tight, mathspacing=tight]{savetrees}
\usepackage{tikz}
\usepackage{xcolor}

\ifx\IsPreprint\undefined
\usepackage{spconf}
\else
\usepackage[preprint]{spconf}

\copyrightnotice{%
\begin{minipage}{\columnwidth}%
\vspace{\baselineskip}%
\footnotesize%
\copyright\ IEEE 2021. Personal use of this material is permitted. Permission from IEEE must be obtained for all other uses, in any current or future media, including reprinting/republishing this material for advertising or promotional purposes, creating new collective works, for resale or redistribution to servers or lists, or reuse of any copyrighted component of this work in other works.
\end{minipage}%
}

\toappear{%
\begin{minipage}{\columnwidth}%
\footnotesize To appear in {\it Proc.\ ICASSP 2021, June 6-11, 2021, Toronto, Canada}%
\vspace*{3em}%
\end{minipage}%
}
\fi

\newcommand{\csection}[1]{
    \section{#1}
}

\newcommand{\csubsection}[1]{
    \vspace{-2mm}
    \subsection{#1}
    \vspace{-2mm}
}

\title{Error-driven Pruning of Language Models for Virtual Assistants}

\name{%
\begin{tabular}{c}%
Sashank Gondala\thanks{$^1$Work done while the first author was an intern at Apple.}\thanks{$^{\star}$Equal contribution.}$^{1,\star}$ \qquad Lyan Verwimp$^{2,\star}$ \qquad Ernest Pusateri$^2$ \\
Manos Tsagkias$^2$ \qquad Christophe Van Gysel$^2$%
\end{tabular}}
\address{
    \textsuperscript{$^1$}Georgia Institute of Technology, \textsuperscript{$^2$}Apple\\
    \texttt{\href{mailto:sgondala@gatech.edu}{sgondala@gatech.edu}},\\
    \texttt{\{\href{mailto:lverwimp@apple.com}{lverwimp},
              \href{mailto:epusateri@apple.com}{epusateri},
              \href{mailto:etsagkias@apple.com}{etsagkias},
              \href{mailto:cvangysel@apple.com}{cvangysel}\}@apple.com}
}

\begin{document}

\newcommand{\ngram}{n-gram}
\newcommand{\ngrams}{\ngram s}
\newcommand{\Ngram}{N-gram}
\newcommand{\Ngrams}{\Ngram s}

\newcommand{\KeepList}{keep list}

\newcommand{\RegularPruning}{EP}
\newcommand{\DomainDependentPruning}{QEP}
\newcommand{\ErrorDrivenPruning}{EEP}
\newcommand{\ApproximateErrorDrivenPruning}{AEEP}

\newcommand{\TTSASRLoop}{TTS--ASR loop}

\newcommand{\Query}{q}
\newcommand{\QuerySet}{\MakeUppercase{\Query{}}}
\newcommand{\Length}[1]{|#1|}

\maketitle

\begin{abstract}
Language models (LMs) for virtual assistants (VAs) are typically trained on large amounts of data, resulting in prohibitively large models which require excessive memory and/or cannot be used to serve user requests in real-time. Entropy pruning results in smaller models but with significant degradation of effectiveness in the tail of the user request distribution. We customize entropy pruning by allowing for a keep list of infrequent n-grams that require a more relaxed pruning threshold, and propose three methods to construct the keep list. Each method has its own advantages and disadvantages with respect to LM size, ASR accuracy and cost of constructing the keep list.
Our best LM gives 8\% average Word Error Rate (WER) reduction on a targeted test set, but is 3 times larger than the baseline. We also propose discriminative methods to reduce the size of the LM while retaining the majority of the WER gains achieved by the largest LM.
\end{abstract}

\csection{Introduction}
\label{section:introduction}
\vspace{-2mm}

VAs are popular services \cite{Juniper2019popularity} that help users accomplish multiple tasks through voice queries. The Automatic Speech Recognition (ASR) engine, the VA component responsible for converting spoken queries into text, faces a challenge due to the many task domains VAs supports. Task domains include performing actions on the device where the VA runs (e.g. placing a call on a cell phone) or querying information about real-world events such as the outcome of a sports competition.

Accurately recognizing queries that concern contemporary events is a difficult problem due to the dynamic nature of the world. Hence, within VA systems, the language models (LMs) of the ASR system are typically trained on synthetic queries that are generated from knowledge bases \cite{Gandhe2018sds}, in addition to transcribed user queries. For example, when the artist Kanye West announced his album \emph{Donda: With Child} in July 2020, the artificial queries \emph{``play Donda With Child''} and \emph{``what is Donda With Child''} are included within LM training data.  %
Likewise, artificial queries corresponding to entities that may occur infrequently within real usage data (i.e., tail entities), but should still be recognized accurately, are also included within LM training data.

As users expect low-latency responses from online services \cite{Schurman2009latency}, {\ngram} backoff LMs~\cite{Katz1987backoff} are frequently used, along with entropy-based pruning \cite{Stolcke1998entropy} to reduce speech recognition runtime and memory consumption. %
Entropy-based pruning removes {\ngrams} from an LM that have the smallest impact on training set perplexity. This can be problematic if explicitly-observed {\ngrams} from synthetic queries are removed, making them indistinguishable from unobserved {\ngrams} in the training data. %
However, entropy pruning can be modified to apply different pruning thresholds to a subset of LM {\ngrams}, and hence, we use a more relaxed pruning threshold for specific (tail) {\ngrams} that enhance recognition.

In this paper we investigate how to determine the minimal set of {\ngrams} that require a more relaxed pruning threshold from textual features.
In particular, we are interested in improving recognition on a set of synthetic queries $\QuerySet{}$ that we know we want our VA to recognize, but are absent or underrepresented in our live usage data. This problem is challenging due to the large number of synthetic queries, the lack of generalization in {\ngram} LMs, the dynamicity of live usage data, and VA runtime constraints.%
\footnote{We also tried a variety of different smoothing methods \cite{Gao2002pruning} to try to achieve the same goal, including Kneser-Ney smoothing \cite{Chen1999smoothing, Chelba2010entropy}; no significant improvements on our tail- and entity-rich test sets were observed.}

Our research questions are:
\begin{enumerate*}[label=(\textbf{RQ\arabic*})]
    \item\label{rq:1} Are there text-based signals that are a good predictor for speech recognition difficulty?
    \item\label{rq:2} Can we determine an optimal subset of synthetic query {\ngrams} that need a more relaxed pruning threshold without degrading speech recognition effectiveness? 
\end{enumerate*}
We contribute: %
\begin{enumerate*}[label=(\arabic*)]
    \item a formal framework for applying different entropy-based pruning thresholds on subsets of {\ngrams},
    \item three methods for determining a subset of {\ngrams} for which pruning needs to be relaxed to improve their recognition,
    \item insight into which signals are useful to predict speech recognition difficulty directly from text.
\end{enumerate*}
\csection{Customized pruning for queries}
\label{section:pruningmethods}
\vspace{-2mm}

We describe three methods for pruning backoff LMs, all of which build upon entropy-based pruning \cite{Stolcke1998entropy}, with the goal of improving recognition on a set of synthetic queries $\QuerySet{}$.  %
In entropy-based pruning (\textit{\RegularPruning{}}), \ngrams{} that are estimated to increase training data perplexity by less than a threshold, $\theta$, are greedily removed from a backoff \ngram{} LM.  We extend this approach by introducing a \emph{\KeepList{}} of \ngrams{}.  The \KeepList{} defines a set of n-grams, extracted from the queries in $\QuerySet{}$, for which a more relaxed threshold, $\theta_\text{keep}$, will be applied. Notice that even though we apply this extension to entropy pruning and a set of synthetic queries, it can in principle be applied to any type of \ngram{} LM pruning (e.g.~\cite{Seymore1996scalable}) and any dataset for which you want to improve recognition.  %
The three methods described below differ only in the way the \KeepList{} is generated.  Each method provides a different space of trade-offs between \KeepList{} generation cost, LM size, and ASR accuracy.  %

\csubsection{Query-driven Entropy Pruning}
\label{section:modeling-custom}

In query-driven entropy pruning (\textit{\DomainDependentPruning{}}), we generate the \KeepList{} by extracting \ngrams{} from every query $\Query{} \in \QuerySet{}$. Using the full set of synthetic queries does not require extra processing and gives the best ASR accuracy on tail- and entity-rich test sets, but can easily blow up the size of the LM. While effective, \DomainDependentPruning{} is infeasible to apply when $\QuerySet{}$ is large and contains many unique n-grams.

\csubsection{Error-driven Entropy Pruning}
\label{section:modeling-discr}

To reduce the number of \ngrams{} in the \KeepList{}, we apply a discriminative approach (e.g.,~\cite{Roark2004corrective,Oba2012duel,Tachioka2015discriminative}), error-driven entropy pruning (\textit{\ErrorDrivenPruning{}}).  Here we only want to exclude those \ngrams{} from pruning for which our baseline ASR, using an entropy-pruned LM, fails.
However, large-scale manual transcription of audio is expensive, and we want to optimize the ASR accuracy of data that is possibly heavily underrepresented in real user data.  %
Therefore, for every phrase in $Q$, we generate audio with Text-To-Speech (TTS) and recognize it with our baseline ASR (see~\cite{Guo2019spelling} for a similar approach). The decoding errors are then used as a source to extract \ngrams{} for the \KeepList{}. This \TTSASRLoop{} is costly both in terms of time and computational resources and thus prohibitive for large datasets.

\csubsection{Approximate Error-driven Entropy Pruning}
\label{section:modeling-approx}

\newcommand{\TrainingSet}{T_\QuerySet{}}

The \TTSASRLoop{} can be avoided if we can predict, based on textual features alone, whether a query $q$ will be recognized incorrectly by the baseline ASR. Approximate error-driven entropy pruning (\textit{\ApproximateErrorDrivenPruning{}}) -- works as follows: we train a binary classifier (see \S\ref{section:aed-results}) on a training set $\TrainingSet{}$, which is sampled from the same distribution as $\QuerySet{}$, that predicts the outcome of the \TTSASRLoop{}, and more specifically, whether an \ngram{} will be recognized incorrectly. After training, we use that classifier to select a subset of $\QuerySet{}$ as keep list.

For every unique query $\Query{}_i \in \TrainingSet{}$ in our dataset, we apply the \TTSASRLoop{} to obtain $\widetilde{\Query{}_i}$, the top-hypothesis after recognizing the synthesized audio of phrase $\Query{}_i$. 
Subsequently, we obtain an alignment (edit distance) between $\Query{}_i$ and $\widetilde{\Query{}}_i$ and extract {\ngrams} from $\Query{}_i$. The input to our model is an {\ngram} extracted from $\Query{}_i$ and it is assigned a positive label if the target token in the {\ngram} (e.g.\ the target token in a 4-gram is the 4th token) differs from its aligned token in $\widetilde{\Query{}}_i$.
Every instance is represented as a real-valued feature vector. We consider five categories of features:  %

\noindent\textbf{Word-level features.} We compute \ngram{} statistics, including \ngram{} count and frequency in $\TrainingSet{}$, and whether the \ngram{} context words and/or target word are out-of-vocabulary.

\noindent\textbf{Language model features.} We compute features derived from the baseline entropy-pruned LM, such as log probability, perplexity and entropy, of the full \ngram{} and of the target word given the context.

\noindent\textbf{Phoneme-level features.} Our VA lexicon is generated based on a list of pre-defined word-pronunciation mappings for frequent words and exceptions, and a grapheme-to-phoneme tool (G2P, see \S\ref{section:setup}) that automatically generates pronunciations for words that are not in that list. For a given \ngram{}, we generate phonemes for each word individually with our G2P (even when the word occurs in the pre-defined list) and join them together to create the phoneme string and corresponding phoneme \ngrams{}.
From these, we extract the following features: the number of phonemes, whether the phoneme string contains infrequent phonemes, whether the word-pronunciation mapping occurs in the pre-defined list, and the edit distance between the phoneme string generated using G2P and the pre-defined pronunciation (if present).

\noindent\textbf{Template features.} Our dataset of synthetic queries consists of a set of templates (e.g. ``Who is the \verb|position-tag| of \verb|team-tag|?'') where the tags are replaced with entities.  
We tag every word of the utterance to be either a template token or entity and create two features: whether the target word is an entity and whether any word in the \ngram{} is an entity. 

\noindent\textbf{Phonetic confusion features.} For each \ngram{}, we take the target word, extract its phoneme string and compute the set of phoneme strings with an edit distance of 1. For each phoneme string in that set, we check if it maps to any of the words in the vocabulary and if it does, we compute the total log probability of the \ngram{} where the target word is replaced with the new mapped word. We store the highest of these log probabilities as a feature.

\csection{Experimental setup}
\label{section:setup}
\vspace{-2mm}

\noindent\textbf{Training LMs.} We train our LMs on manually and automatically transcribed anonymized VA requests, and on the dataset of synthetic queries $\QuerySet{}$, which consists of a large set of domains, e.g. sports, music and home automation. The queries consist of a set of templates with slots and a list of entities that can fill those slots~(\S\ref{section:modeling-approx}). Both templates and entities are typically derived from real user data and thus have have prior probabilities that are used to sample the synthetic requests. In our experiments, query set $\QuerySet{}$ consists of 2.8B queries (112M unique) spanning 27 domains. The total training data contains more than 10B utterances.\footnote{We are unable to provide the exact number due to confidentiality. The manually transcribed dataset is a small random sample.}

\noindent\textbf{Pruning methods.} Query set $\QuerySet{}$ is also used for our three proposed pruning methods to extract the following: all possible {\ngrams} (\DomainDependentPruning{}), a subset of {\ngrams} based on the errors collected through the {\TTSASRLoop{}} (\ErrorDrivenPruning{}), and a subset of {\ngrams} based on the classifier trained on the output of the {\TTSASRLoop{}} (\ApproximateErrorDrivenPruning{}). In addition, we compare against regular entropy pruning (\RegularPruning{}), where $\QuerySet{}$ is not taken into account.

\noindent\textbf{Evaluating the \ApproximateErrorDrivenPruning{} classifier.} 
To assess the utility of \ApproximateErrorDrivenPruning{}, we train on all domains except one and validate and test on a held-out domain. We consider only one held out domain, that of sports, because it is the only domain for which we have domain-specific test sets using both manually transcribed user requests and synthesized requests. %

\noindent\textbf{ASR accuracy evaluation.} %
After applying the pruning methods outlined above, we evaluate WER on three main test sets:
\begin{enumerate*}[label=(\arabic*)]
    \item General VA contains VA requests sampled from the actual distribution, thus containing mostly frequent/head utterances.
    \item Sports is a sample of less frequent VA requests for the sports domain.
    \item TTS-All consists of synthesized requests, it is sampled from the same templates as $\QuerySet{}$ but is a different instance. Thanks to the prior probabilities, we can make a distinction between subsets of TTS-All that contain utterances from the head (top 10\%), torso (10-50\%) and tail (50-100\%) of the distribution. TTS-Sports is a subset of TTS-All containing all utterances related to sports.
\end{enumerate*} %
Our goal is to improve WER (with a minimum increase in LM size) on the Sports and TTS-All test sets, while not degrading recognition effectiveness on the General VA test set.

\noindent\textbf{System description.}
\label{section:system}
Our ASR system consists of an acoustic model that is a deep convolutional neural network~\cite{Huang2020sndcnn}, a 4-gram LM with Good-Turing smoothing in the first pass (see \cite{Pusateri2019interpolation,VanGysel2020entities} for details), and the same LM interpolated with a Feed-Forward Neural Network (FFNN) LM~\cite{Zhang2015fofe} in the second pass. 
To build a scalable \TTSASRLoop{}, we use our previous generation speech synthesizer, a unit selection system described in~\cite{Capes2017tts}. We use scikit-learn~\cite{Pedregosa2011scikit} to train binary classifiers for \ApproximateErrorDrivenPruning{}. For combining entropy pruning with a keep list we modified SRILM~\cite{Stolcke2020srilm}. Our G2P is an LSTM encoder-decoder architecture with attention (similar to \cite{Toshniwal2016g2p}).
For all experiments we set $\theta=6e-9$ and $\theta_{keep}=0$, which means that all {\ngrams} in the keep list are excluded from pruning. We experimented with different values for $\theta_{keep}$ as an alternative approach to reduce the size of the LM, but did not observe good improvements.
All experiments reported here are for American English.\footnote{For \DomainDependentPruning{} compared to \RegularPruning{}, we observe similar WER reductions for several other languages/regional variants, e.g., German and Mandarin Chinese.}

\begin{table*}[t]
    \caption{WER results (best result per test set in bold) and size (last row) for LMs with various pruning strategies. \RegularPruning{} base is the baseline, \RegularPruning{}$\sim$ LMs are entropy-pruned LMs of the same size as the proposed models. \ApproximateErrorDrivenPruning{} uses all features, and \ApproximateErrorDrivenPruning{}--Top-3 uses the top-3 features after feature selection. `All domains' and `Sports domain' in the top row refer to the query set used to generate the \ngram{} set.}\label{tab:wer}
    \centering\small
    \begin{tabular}{ l r | r | r r r r | r r r r r r}
        \toprule
        \multirow{2}{*}{\textbf{Test set}}
        &\multirow{2}{*}{\textbf{\# utts.}}
        &\multicolumn{1}{|r}{\textbf{\RegularPruning{}}}
        &\multicolumn{4}{|c}{\textbf{All domains}}   
        &\multicolumn{6}{|c}{\textbf{Sports domain}} \\
        
        &&\textbf{base}
        &\multicolumn{1}{r}{\textbf{\RegularPruning{}$\sim$}}
        &\multicolumn{1}{r}{\textbf{\DomainDependentPruning{}}}
        &\multicolumn{1}{r}{\textbf{\RegularPruning{}$\sim$}}
        &\multicolumn{1}{r}{\textbf{\ErrorDrivenPruning{}}}
        &\multicolumn{1}{|r}{\textbf{\RegularPruning{}$\sim$}}
        &\multicolumn{1}{r}{\textbf{\DomainDependentPruning{}}}
        &\multicolumn{1}{r}{\textbf{\RegularPruning{}$\sim$}}
        &\multicolumn{1}{r}{\textbf{\ErrorDrivenPruning{}}}
        &\multicolumn{1}{r}{\textbf{\ApproximateErrorDrivenPruning{}}}
        &\multicolumn{1}{r}{\textbf{\ApproximateErrorDrivenPruning{}--Top-3}} \\
        
        \midrule
        General VA  &49k&4.13
        &\textbf{4.08}&4.13&4.09&4.13
        &4.13&4.13&4.14&4.13&4.13&4.13 \\
        Sports  &4k&4.72
        &4.47&4.32&4.44&4.41
        &4.70&\textbf{4.25}&4.69&4.38&4.27&4.31 \\
        TTS-All &100k&15.07
        &14.42&\textbf{13.89}&14.64&14.00
        &14.95&14.22&15.00&14.27&14.26&14.27 \\
        \quad Head  &32k&8.54
        &7.85&\textbf{7.38}&8.05&7.49
        &8.40&7.73&8.46&7.76&7.78&7.78 \\
        \quad Torso &34k&15.38
        &14.64&\textbf{13.79}&14.91&13.95
        &15.23&14.15&15.29&14.24&14.22&14.24 \\
        \quad Tail  &34k&20.80
        &20.25&\textbf{19.99}&20.47&20.06
        &20.71&20.28&20.76&20.31&20.29&20.29 \\
        \quad TTS-Sports    &28k&19.12
        &18.37&15.82&18.64&15.95
        &18.98&\textbf{15.75}&19.04&15.95&15.93&15.96 \\
        \midrule
        \multicolumn{2}{l|}{\textbf{Number of \ngrams{}}}&22M
        &71M&71M&45M&45M
        &27M&27M&25M&25M&25M&25M \\
        \bottomrule
    \end{tabular}
\vspace{-4mm}
\end{table*}

\csection{Results}
\label{section:results}
\vspace{-2mm}

\newcommand{\RQRef}[1]{\hyperref[rq:#1]{\textbf{RQ#1}}}

\subsection{\ApproximateErrorDrivenPruning{} Classifier}
\label{section:aed-results}

\begin{figure}[t]
\centering
\resizebox{\columnwidth}{!}{\begin{tikzpicture}
\begin{axis}[
    xmin=0, xmax=100,
    xtick=data,
    ymin=0, ymax=1.05,
    ytick={0.0, 0.20, 0.40, 0.60, 0.80, 1.00},
    xlabel={Percentage of {\ngrams} assigned to positive class/keep list ($k$)},
    every axis y label/.style={at={(ticklabel cs:0.5)},rotate=90,anchor=center,yshift=5pt},
    ylabel={Recall@$k$},
    every axis plot/.append style={ultra thick},
    legend style={at={(0.75,0.40)},anchor=north,font=\footnotesize},
    width=8cm,
    height=4.5cm
]
\addplot[cyan,mark=*] table {images/cross-domain_sports.dat};
\addlegendentry{Sports domain};
\addplot[red,dashed,mark=circle*] table {images/equal.dat};
\addlegendentry{Y = X};
\end{axis}
\end{tikzpicture}}
\setlength{\abovecaptionskip}{-5pt}
\setlength{\belowcaptionskip}{-10pt}
\vspace{-7mm}
\caption{Recall@$k$ for the sports domain.}
\vspace{-5mm}
\label{fig:recall_at_k}
\end{figure}
\vspace{-2mm}
In this section, we report the results for training a binary classifier on the output of the \TTSASRLoop{} to predict whether an {\ngram} will be recognized incorrectly.
Comparing 4 different model types, Random Forests (RF), AdaBoost, linear support-vector machines and FFNNs, we found that RFs achieved the best results on the validation data. Hence, all experiments reported below use RFs.  

We parameterize our experiments based on the number of {\ngrams} to consider: We take the top-$k$\% of {\ngrams} as ranked by the classifier confidence score and assign them to the positive class that will form the keep list.
In Fig.~\ref{fig:recall_at_k}, we show recall@$k$ for the sports domain: the relative number of {\ngrams} correctly classified as positive if $k$\% of the ranked data is assigned to the positive class/keep list. We observe good performance, with a recall of 0.65 at 40\%. For the downstream task (\S\ref{section:asr-results}), we select this 40\% of \ngrams{} as the keep list for \ApproximateErrorDrivenPruning{} (Sports).
For our ablation study, we rank the features according to the importance assigned to them by the RF. We find that by using just the top-3 features to train a new RF, i.e.\
\begin{enumerate*}[label=(\arabic*)]
    \item log probability of target word given context,
    \item log probability of the full {\ngram} and
    \item largest log probability of the {\ngram} at an edit distance of 1,
\end{enumerate*}
we still obtain a recall of 0.63 at 40\%. We will use the data selected by this model as well in \S\ref{section:asr-results} (\ApproximateErrorDrivenPruning{}--Top-3).

These results provide an answer to \RQRef{1}: Yes, there are text-based signals that can predict ASR difficulty. LM and phonetic confusion features are the most important ones.

\csubsection{ASR Accuracy}
\label{section:asr-results}

We now evaluate our 3 customized pruning approaches in our ASR system. We set our success criterion to minimize the size of a LM while improving WER compared to that of a system using an entropy-pruned LM. In the last row of Table~\ref{tab:wer}, we show the size of each LM in number of \ngrams{}. \DomainDependentPruning{} roughly triples the size of the entropy-pruned LM, while \ErrorDrivenPruning{} doubles it. The unpruned LM contains 313M {\ngrams} and is too large to use in ASR decoding. In our ASR system, using the larger LMs leads to increased, but still acceptable, memory usage, and we observe negligible impact on the decoding speed. Since all our proposed approaches increase the size of the LM, we created additional \RegularPruning{} baselines of the same size for each pruning strategy for a fair comparison in terms of LM size -- see the \RegularPruning{}$\sim$ columns in Table~\ref{tab:wer}.

The first row of Table~\ref{tab:wer} shows that none of the pruning approaches hurts the WER on our regular, head-heavy general VA test set. 
On the test sets representing the usage that we want to improve on (i.e., tail utterances with many named entities), we see consistent WER reductions for all settings. 

\DomainDependentPruning{} leads to the largest LMs and the best WER results on TTS-All, with 8\% relative WER reduction compared to the baseline \RegularPruning{} LM, while the \RegularPruning{} LM of the same size (\RegularPruning{}$\sim$ column to the left of \DomainDependentPruning{}) has only a 4\% reduction.
On the sports test with user queries we also observe a WER reduction of 8\%, compared to 5\% for \RegularPruning{}$\sim$. Looking at results qualitatively, we observe that the \RegularPruning{} LMs do not contain many higher order {\ngrams} that make a difference, e.g.\ \textit{who is Phil Bengtson} (in TTS-All) is recognized correctly by the \DomainDependentPruning{} LM because it contains all relevant higher-order {\ngrams} while the \RegularPruning{} LMs only contain the unigrams.
  
The \ErrorDrivenPruning{} LM, that is significantly smaller, only leads to small WER degradations compared to the \DomainDependentPruning{} LM. We observe 7\% relative WER reduction w.r.t to the baseline on TTS-All, compared to only 3\% for the corresponding \RegularPruning{}$\sim$ LM.

Finally, approximating the \TTSASRLoop{} with \ApproximateErrorDrivenPruning{} gives us WER results that are close to the real decoding errors (\ErrorDrivenPruning{}). The \ApproximateErrorDrivenPruning{} model using all features described in \S\ref{section:modeling-approx} gives a 17\% relative WER reduction with respect to the baseline \RegularPruning{} LM on the TTS-Sports test set, 
and 10\% on the Sports test set extracted from real user data. \ApproximateErrorDrivenPruning{}--Top-3, the model using only the top-3 features (\S\ref{section:aed-results}), is on par with the variant using all features.
On the Sports test set with real user requests, \ApproximateErrorDrivenPruning{} does even better than \ErrorDrivenPruning{}. We hypothesize that \ErrorDrivenPruning{} is overfitting on the data that is used for the \TTSASRLoop{}, while \ApproximateErrorDrivenPruning{} counteracts this problem because it approximates the results of the \TTSASRLoop{}.

For both the TTS-based error-driven pruning (\ErrorDrivenPruning{}) and its approximation (\ApproximateErrorDrivenPruning{}), we select about 40\% of the full Sports dataset for the keep list. One could argue that any addition of sports-related {\ngrams} to the LM will improve the WER on sports test sets. Thus, as a sanity check we randomly select 40\% of the sports {\ngrams} and use it as the keep list. The resulting LM gives a WER of 16.27 on the TTS-Sports test set, which is still 0.3 absolute worse than \ApproximateErrorDrivenPruning{}. We can conclude that the 40\% selected by \ErrorDrivenPruning{} and \ApproximateErrorDrivenPruning{} is a more meaningful selection than a randomly selected 40\%.

We can conclude that w.r.t. \RQRef{2} the optimal set of synthetic query {\ngrams} that require a more relaxed pruning threshold can be the full set (\S\ref{section:modeling-custom}) if there are no memory limits. However, we showed that it is possible to retain the majority of the WER gain with a much smaller LM by selecting {\ngrams} with a model trained on textual features alone.

\vspace{-1mm}
\csection{Conclusions}
\label{section:conslusions}
\vspace{-2mm}
We explored three methods to customize LM pruning to improve ASR accuracy on infrequent and entity-rich utterances, by constructing a keep list of {\ngrams} that require a more relaxed pruning threshold. \DomainDependentPruning{} results in LMs that are three times larger than the baseline and give relative WER reductions of on average 8\%, both on a targeted synthetic test set and a test set with user queries. \ErrorDrivenPruning{} and the more efficient \ApproximateErrorDrivenPruning{} reduce the size of the keep list by selecting only decoding errors, resulting in LMs that are only twice as large as the baseline and still have good WER improvements of 17\% on the domain-specific synthetic test set and 10\% on the user query test set.
We also showed that we can predict ASR difficulty based on textual signals.

In our future work, we would like to explore more text-only approaches to customize pruning, e.g.\ by modifying the pruning criterion itself or selecting data based on LM features and improved (FST-based) phonetic confusion features.

\vspace{1em}
\begin{small}
\noindent {\bf Acknowledgements.} We thank Youssef Oualil, Amr Mousa, Russ Webb, Barry Theobald for their comments and feedback.
\end{small}

\bibliographystyle{IEEEbib}
\small%
\bibliography{icassp2021-dalec}

\begin{thebibliography}{10}

\bibitem{Juniper2019popularity}
{Juniper Research},
\newblock ``{Digital Voice Assistants in Use to Triple to 8 Billion by 2023,
  Driven by Smart Home Devices},'' Press Release, Feb. 2019.

\bibitem{Gandhe2018sds}
Ankur Gandhe, Ariya Rastrow, and Bjorn Hoffmeister,
\newblock ``Scalable language model adaptation for spoken dialogue systems,''
\newblock in {\em IEEE Spoken Language Technology Workshop (SLT)}, 2018, pp.
  907--912.

\bibitem{Schurman2009latency}
Eric Schurman and Jake Brutlag,
\newblock ``{Performance Related Changes and their User Impact},'' Presentation
  at Velocity -- Web Performance and Operations Conference, 2009.

\bibitem{Katz1987backoff}
Slava Katz,
\newblock ``Estimation of probabilities from sparse data for the language model
  component of a speech recognizer,''
\newblock {\em Transactions on Acoustics, Speech, and Signal Processing}, vol.
  35, no. 3, pp. 400--401, 1987.

\bibitem{Stolcke1998entropy}
Andreas Stolcke,
\newblock ``{Entropy-based Pruning of Backoff Language Models},''
\newblock {\em ArXiv}, vol. cs.CL/0006025, 1998.

\bibitem{Gao2002pruning}
Jianfeng Gao and Min Zhang,
\newblock ``{Improving Language Model Size Reduction using Better Pruning
  Criteria},''
\newblock in {\em Proceedings of the 40th Annual Meeting of the Association for
  Computational Linguistics (ACL)}, 2002, pp. 176--182.

\bibitem{Chen1999smoothing}
Stanley~F. Chen and Joshua Goodman,
\newblock ``{An empirical study of smoothing techniques for language
  modeling},''
\newblock {\em Computer Speech and Language}, vol. 13, pp. 359--394, 1999.

\bibitem{Chelba2010entropy}
Ciprian Chelba, Thorsten Brants, Will Neveitt, and Peng Xu,
\newblock ``{Study on Interaction between Entropy Pruning and Kneser-Ney
  Smoothing},''
\newblock in {\em Proceedings Interspeech}, 2010, pp. 2242--2245.

\bibitem{Seymore1996scalable}
Kristie Seymore and Ronald Rosenfeld,
\newblock ``{Scalable backoff language models},''
\newblock in {\em International Conference on Spoken Language Processing
  (ICSLP)}, 1996, pp. 232--235.

\bibitem{Roark2004corrective}
Brian Roark, Murat Saraclar, and Michael Collins,
\newblock ``{Corrective language modeling for large vocabulary ASR with the
  perceptron algorithm},''
\newblock in {\em IEEE International Conference on Acoustics, Speech and Signal
  Processing (ICASSP)}, 2004, pp. 749--752.

\bibitem{Oba2012duel}
Takanobul Oba, Takaaki Hori, Atsushi Nakamura, and Akinori Ito,
\newblock ``{Round-Robin Duel Discriminative Language Models},''
\newblock {\em Transactions on Audio, Speech, and Language Processing}, vol.
  20, no. 4, pp. 1244--1255, 2012.

\bibitem{Tachioka2015discriminative}
Yuuki Tachioka and Shinji Watanabe,
\newblock ``Discriminative method for recurrent neural network language
  models,''
\newblock in {\em IEEE International Conference on Acoustics, Speech and Signal
  Processing (ICASSP)}, 2015, pp. 5386--5390.

\bibitem{Guo2019spelling}
Jinxi Guo, Tara~N. Sainath, and Ron~J. Weiss,
\newblock ``{A Spelling Correction Model for End-to-end Speech Recognition},''
\newblock in {\em IEEE International Conference on Acoustics, Speech and Signal
  Processing (ICASSP)}, 2019, pp. 5651--5655.

\bibitem{Huang2020sndcnn}
Zhen Huang, Tim Ng, Leo Liu, Henry Mason, Xiaodan Zhuang, and Daben Liu,
\newblock ``{SNDCNN: Self-Normalizing Deep CNNs with Scaled Exponential Linear
  Units for Speech Recognition},''
\newblock in {\em IEEE International Conference on Acoustics, Speech and Signal
  Processing (ICASSP)}, 2020, pp. 6854--6858.

\bibitem{Pusateri2019interpolation}
Ernest Pusateri, Christophe~Van Gysel, Rami Botros, Sameer Badaskar, Mirko
  Hannemann, Youssef Oualil, and Ilya Oparin,
\newblock ``{Connecting and Comparing Language Model Interpolation
  Techniques},''
\newblock in {\em Proceedings Interspeech}, 2019, pp. 3500--3504.

\bibitem{VanGysel2020entities}
Christophe Van~Gysel, Manos Tsagkias, Ernest Pusateri, and Ilya Oparin,
\newblock ``{Predicting Entity Popularity to Improve Spoken Entity Recognition
  by Virtual Assistants},''
\newblock in {\em SIGIR}, 2020, pp. 1613--1616.

\bibitem{Zhang2015fofe}
Shiliang Zhang, Hui Jiang, Mingbin Xu, Junfeng Hou, and Lirong Dai,
\newblock ``{The Fixed-Size Ordinally-Forgetting Encoding Method for Neural
  Network Language Models},''
\newblock in {\em IJCNLP}, 2015, pp. 495--500.

\bibitem{Capes2017tts}
Tim Capes, Paul Coles, Alistair Conkie, Ladan Golipour, Abie Hadjitarkhani,
  Qiong Hu, Nancy Huddleston, Melvyn Hunt, Jiangchuan Li, Matthias Neeracher,
  Kishore Prahallad, Tuomo Raitio, Ramya Rasipuram, Greg Townsend, Becci
  Williamson, David Winarsky, Zhizheng Wu, and Hepeng Zhang,
\newblock ``{Siri On-Device Deep Learning-Guided Unit Selection Text-to-Speech
  System},''
\newblock in {\em Proceedings Interspeech}, 2017, pp. 4011--4015.

\bibitem{Pedregosa2011scikit}
Fabian Pedregosa, Ga{\"e}l Varoquaux, Alexandre Gramfort, Vincent Michel,
  Bertrand Thirion, Olivier Grisel, Mathieu Blondel, Peter Prettenhofer, Ron
  Weiss, Vincent Dubourg, et~al.,
\newblock ``{Scikit-learn: Machine learning in Python},''
\newblock {\em Journal of machine learning research}, vol. 12, pp. 2825--2830,
  2011.

\bibitem{Stolcke2020srilm}
Andreas Stolcke,
\newblock ``{SRILM -- An extensible language modeling toolkit},''
\newblock in {\em International Conference on Spoken Language Processing
  (ICSLP)}, 2002, pp. 901--904.

\bibitem{Toshniwal2016g2p}
Shubham Toshniwal and Karen Livescu,
\newblock ``Jointly learning to align and convert graphemes to phonemes with
  neural attention models,''
\newblock in {\em IEEE Spoken Language Technology Workshop (SLT)}, 2016, pp.
  76--82.

\end{thebibliography}

\end{document}